\documentclass[11pt]{article}

\usepackage[a4paper,margin=1in]{geometry}
\usepackage{microtype}
\usepackage[section]{placeins}


\usepackage{amsmath,amssymb,amsthm}

\usepackage{graphicx}
\graphicspath{{figures/}{./}}
\usepackage{booktabs}
\usepackage[font=small,labelfont=bf]{caption}
\newtheorem{corollary}{Corollary}
\usepackage{enumitem}
\usepackage{xcolor}
\usepackage[numbers,sort&compress]{natbib}
\usepackage[colorlinks=true,linkcolor=blue,citecolor=blue,urlcolor=blue]{hyperref}
\usepackage[capitalise,noabbrev]{cleveref}   %

\newtheorem{proposition}{Proposition}
\newtheorem{lemma}{Lemma}
\theoremstyle{definition}
\newtheorem{remark}{Remark}

\newcommand{\edjm}{EDJM}
\newcommand{\rjm}{RJM}

\title{When Does Jacobian Fine-Tuning Stabilize Neural PDE Emulators?}
\author{%
  Shreeyam Bangera\thanks{Equal contribution.}\\
  IIT Roorkee\\
  \texttt{shreekant{\char95}sb@ph.iitr.ac.in}
  \and
  Manya Gupta\footnotemark[1]\\
  IIT Roorkee\\
  \texttt{manya{\char95}g@ph.iitr.ac.in}
}
\date{}

\begin{document}
\maketitle

\begin{abstract}
Neural emulators of chaotic PDEs are trained on one-step prediction but deployed autoregressively, where their errors compound. Jacobian-based regularisers aim to curb this growth, but in a chaotic system part of the growth is physical and cannot be trained away. We separate the two sources of this error with a first-order decomposition. We compare the emulator's error with a counterfactual that propagates the same one-step mistakes through the solver's Jacobian, giving an amplification ratio: the factor by which the model's own dynamics amplify its errors. We study 21 configurations across four PDEs, five architectures and two model sizes (8--10k and 33k parameters), with errors measured at one Lyapunov time. Jacobian fine-tuning reduced the error by more than 10\% for 1 of 34 seeds whose amplification ratio was at most 1.25, and for all 20 seeds whose ratio was at least four; for the latter, the error fell by a factor of 8 (small models) and 55 (33k models) in geometric mean over seeds. The gain approximately equals the one-step accuracy cost of fine-tuning divided by the amplification ratio. Matching the solver's Jacobian along the model's own error direction (EDJM) outperformed random directions at the same cost on 28 of 35 seeds; input-noise injection performed comparably to EDJM.
\end{abstract}

\section{Introduction}\label{sec:introduction}
In recent years neural networks have been trained to replace numerical solvers across a range of
physics problems: medium-range weather forecasting \citep{lam2023graphcast, bi2023pangu},
turbulent fluid flow \citep{kochkov2021cfd}, climate simulation \citep{wattmeyer2023ace}, and
subsurface flow \citep{wen2022ufno}. A single run can require millions of solver steps, and the
cost of each is what makes high-fidelity simulation expensive; a neural network that emulates one step
can do it faster by orders of magnitude.

These models are trained with a Markovian bias, in the sense that they are only asked to predict the next state given the previous one, and they become accurate on this task. During inference, however, they are deployed \textbf{autoregressively}: each prediction becomes the input to the next call, for hundreds or thousands of steps. Training never presents this situation: the model learns from real simulation data, then runs on inputs it generated
itself, which are slightly wrong after the first step and worse thereafter. Errors compound until
the trajectory bears no relation to the truth --- most severely for chaotic systems, where nearby states separate exponentially at a rate set by
the leading Lyapunov exponent \citep{eckmann1985ergodic, edson2019lyapunov}. Among the models we train here, a Kuramoto--Sivashinsky emulator with a one-step error of 0.4\% loses its forecast skill (normalised error above 0.5) within four steps.

A large body of work aims to slow this growth, through modified training
objectives \citep{brandstetter2022mppde, sanchezgonzalez2020learning,
stachenfeld2022learned} or corrective machinery at inference
\citep{lippe2023pderefiner}. One prominent family instead constrains the
emulator's \textbf{local dynamics}, that is, its Jacobian, the linear map that describes how it responds to small perturbations of its input \citep{mccabe2023stability, nie2026jaws, ainslie2026eigen, kang2026randomized}.
This family assumes that rollout error arises because the emulator responds more violently than the real physics and so manufactures instability the equation does not contain, but it does not check this assumption.
A rollout error has two origins. The system is chaotic, so \emph{any} error grows at a rate fixed by the
equation, and that growth is neither the model's fault nor removable by training. Separately, the model
may carry its own mistakes forward more violently than the true dynamics would --- a genuine
defect these methods could repair. From the outside the two are indistinguishable: a diverging
rollout looks the same either way. Moreover, none of these works reports what the constraint costs in one-step accuracy.

Two of these works already contain the effect. \citet{ainslie2026eigen}
decompose the rollout error into an injected term, set by the one-step
error, and an amplified term, set by the Jacobian; they then observe that
several models trained with their Jacobian-based loss still destabilise
because the injected term remains large. \citet{nie2026jaws} measure the other
side of the trade, reporting a fourfold increase in one-step error
on their out-of-distribution test (0.009 to 0.036) alongside their rollout improvements. But neither evaluates
the two terms against each other for a given model: the first never reports
their relative size, and the second never reports the amplification its
accuracy cost purchased. Neither can therefore establish whether the
intervention was worth making, nor anticipate whether it would be on a
different system.

\paragraph{Contributions.}
\begin{enumerate}[label=\textbf{C\arabic*.}, leftmargin=2.2em]

\item We present a first-order decomposition of the autoregressive error into mistakes propagated by the true dynamics plus a Jacobian-mismatch term. This yields a counterfactual error $e^P_n$: the error of a model with the solver's exact Jacobian that makes the same one-step mistakes.

\item We introduce the amplification ratio $\rho_n=\|e_n\|/\|e^P_n\|$, evaluated at one Lyapunov time, before the error saturates. It separates amplification by the model's Jacobian from the physical instability of the system. Across 21 configurations (over PDEs, architectures and sizes), it identifies when Jacobian matching via fine-tuning can help.

\item We show that $\rho_n$ limits the achievable rollout improvement. Fine-tuning removes the excess Jacobian amplification but changes the one-step accuracy, and the resulting change in rollout error approximately equals this one-step cost divided by $\rho_n$.

\item We propose error-directed Jacobian matching (\edjm{}): matching the solver's Jacobian specifically along the model's own error vector $e_n$. At the same penalty weight and training budget, it outperforms matching along random directions.

\end{enumerate}

\section{Related work}\label{sec:related_work}

\paragraph{Neural emulators and autoregressive rollouts.}
Neural networks emulate one step of a numerical solver across weather forecasting \citep{lam2023graphcast, bi2023pangu, pathak2022fourcastnet}, climate \citep{wattmeyer2023ace}, turbulence \citep{kochkov2021cfd, stachenfeld2022learned} and subsurface flow \citep{wen2022ufno}, using convolutional, graph and operator architectures \citep{li2021fno, lu2021deeponet, kovachki2023neuraloperator, gupta2023multiscale}.
These models are trained on one-step prediction but deployed autoregressively, and their rollouts drift or diverge even when the one-step error is small. Accordingly, benchmarks such as APEBench \citep{koehler2024apebench} report rollout metrics alongside one-step accuracy. Using APEBench architectures on chaotic systems, where error accumulation is intrinsic, we isolate the portion of error amplification (exceeding underlying physical divergence) driven by the model itself.

\paragraph{Training for stable rollouts.}
Most remedies change what the model sees during training.
Input-noise injection \citep{sanchezgonzalez2020learning, stachenfeld2022learned} trains on perturbed states so the model learns to correct small deviations; the pushforward trick and unrolled training \citep{brandstetter2022mppde} expose it to its own predictions; PDE-Refiner \citep{lippe2023pderefiner} adds an iterative refinement step to recover the low-amplitude modes that one-step models neglect; and spectral constraints on the architecture bound the operator norm of each layer \citep{mccabe2023stability}.
These methods are applied uniformly, without first asking whether a given model's instability comes from its own dynamics.
We use input noise and continued training as baselines (\cref{sec:res-arms}). On the small models whose amplification ratio is close to one, neither of them, nor Jacobian matching, improved more than one of 17 seeds (\cref{tab:app-bins}): when a model already propagates its mistakes as the true dynamics would, none of the methods we test helps.

\paragraph{Jacobian and derivative supervision.}
Matching derivatives in addition to values dates back to Sobolev training \citep{czarnecki2017sobolev}.
For PDE surrogates, \citet{behroozi2026scno} supervise the sensitivities of a neural operator with respect to its input parameter fields to improve data efficiency and inverse modelling.
\citet{kang2026randomized} match the Jacobians of the true and learned vector fields at randomly perturbed states, which implicitly enforces second-order consistency and improves the Lyapunov spectra of learned chaotic systems.
Other methods penalise the model's Jacobian without reference to the solver: JAWS \citep{nie2026jaws} penalises its Frobenius norm with a spatially adaptive weight, and \citet{ainslie2026eigen} penalise the one-step residual propagated through the learned Jacobian.
In all of these, the directions along which the Jacobian is constrained are fixed by the data or drawn at random.
\edjm{} instead matches the solver's Jacobian along the direction of the model's own rollout error, the direction in which a Jacobian mismatch actually enters the error (\cref{sec:res-arms}).

\paragraph{Analysing error growth in learned dynamics.}
In a chaotic system any error grows at a rate set by the leading Lyapunov exponent \citep{eckmann1985ergodic, edson2019lyapunov}, so a growing rollout error is not by itself evidence of a defective model.
\citet{ainslie2026eigen} linearise the error dynamics of learned one-step maps and separate an injected one-step error from a term propagated by the model's Jacobian; they use the largest Jacobian eigenvalue as an indicator of error growth.
An eigenvalue above one, however, is expected in a chaotic system and does not distinguish the model's instability from the physics.
We compare the model's error with a counterfactual that propagates the same one-step mistakes through the solver's Jacobian (\cref{sec:res-ledger}).
The resulting amplification ratio measures only the excess amplification that the model adds, identifies when Jacobian fine-tuning can help (\cref{sec:res-diagnostic}), and limits how much it can gain once its cost in one-step accuracy is accounted for (\cref{sec:res-ceiling}).

\section{Method}
\label{sec:method}

In this section, we first define the errors and the counterfactual. We then derive when a Jacobian-matching objective can remove the excess error.

\subsection{Error Notation}
\label{sec:error-notation}
We train a neural emulator $f$ to approximate one time step of the numerical solver $P$. Long-term trajectories are then generated by applying $f$ iteratively to its own outputs.

Let $u_0, u_1, \dots$ be the true trajectory, $u_{n+1} = P(u_n)$. Starting at the same initial state $\hat u_0 = u_0$, the emulator sequence is given by $\hat u_{n+1} = f(\hat u_n)$.
\begin{equation*}
\underbrace{d_n = f(u_n) - P(u_n)}_{\text{the \emph{defect}: one step, from a \emph{correct} state}}
\qquad\qquad
\underbrace{e_n = \hat u_n - u_n}_{\text{the \emph{rollout error}, after $n$ free-running steps}}
\end{equation*}

Training on one-step prediction directly minimizes $d_n$, whereas the rollout error $e_n$ reflects the actual drift experienced in practice.

The defect and rollout error are linked by an exact identity. Since $\hat u_n = u_n + e_n$, we have:
\begin{align}
e_{n+1}
 &= \hat u_{n+1} - u_{n+1} \notag \\
 &= f(\hat u_n) - P(u_n) \notag \\
 &= f(u_n + e_n) - P(u_n) \notag \\
 &= \underbrace{\big[f(u_n + e_n) - f(u_n)\big]}_{\text{propagated error}}
    \;+\; \underbrace{\big[f(u_n) - P(u_n)\big]}_{\text{one-step defect } d_n}
 \label{eq:identity} \\[4pt]
 &= \underbrace{J_f(u_n)\, e_n \;+\; \mathcal{O}\big(\|e_n\|^2\big)}_{\text{Taylor expansion of } f \text{ about } u_n}
    \;+\; d_n \notag \\[4pt]
 &\approx \; J_f(u_n)\, e_n \;+\; d_n . \notag
\end{align}
Equation~\eqref{eq:identity} is exact. Neglecting the $\mathcal{O}(\|e_n\|^2)$ remainder yields the linearized error $e^f_n$, governed by:
\begin{equation}
e^{f}_{n+1} = J_f(u_n)\, e^{f}_n + d_n , \qquad e^{f}_0 = 0 .
\label{eq:linear}
\end{equation}
While the error is small, $e_n = e^f_n + \mathcal{O}(\|e\|^2)$.

\subsection{Counterfactual Notation}
\label{sec:ledger}

Let $J_P = DP$ be the solver's Jacobian. Running the same recursion as \eqref{eq:linear} with $J_P$ in place of
$J_f$, and with the same measured defects, gives
\begin{equation}
e^{P}_{n+1} = J_P(u_n)\, e^{P}_n + d_n , \qquad e^{P}_0 = 0 .
\label{eq:counterfactual}
\end{equation}
$e^P_n$ is the error of a hypothetical emulator that commits \emph{exactly the same
mistakes} but carries them forward with the true physics; we call it the \emph{counterfactual error}.

Comparing the real rollout error to this baseline gives the \textbf{amplification ratio}:
\begin{equation}
\rho_n \;=\; \frac{\|e_n\|}{\|e^{P}_n\|} .
\label{eq:rho}
\end{equation}
We use the actual error $e_n$ so that $\rho_n$ describes the model's real rollouts; to first order it equals $\|e^f_n\|/\|e^P_n\|$, which is the quantity the analysis below uses.
$\rho_n \approx 1$ indicates that the rollout error is just what the defects force under the true dynamics; $\rho_n \gg 1$ indicates that the emulator inflates its own mistakes, and a better Jacobian can remove this excess.
Crucially, $\rho_n$ can be measured \emph{before} any intervention is attempted.

\subsection{Where the Mismatch Acts}
\label{sec:mismatch}

To analyze how Jacobian errors drive trajectory drift, we define two quantities:
\begin{align}
\Delta_n &= J_f(u_n) - J_P(u_n) \quad &&\text{(Jacobian mismatch)} , \\
g_n &= e^{f}_n - e^{P}_n, \qquad g_0 = 0 \quad &&\text{(accumulated error gap)} .
\label{eq:delta}
\end{align}

\begin{lemma}[Error Gap Accumulation]
\label{lem:gap}
Let $\Phi^{P}_{n,m} = J_P(u_{n-1}) \cdots J_P(u_m)$ denote the ground-truth state transition matrix for $n > m$, with $\Phi^{P}_{m,m} = I$. The error gap at step $n$ satisfies:
\begin{equation}
g_n = \sum_{j<n} \Phi^{P}_{n,j+1} \, \Delta_j \, e^{f}_j .
\label{eq:gap}
\end{equation}
\end{lemma}

\begin{proof}
Subtracting $e^{P}_{n+1}$ from $e^{f}_{n+1}$ cancels the one-step defects $d_n$:
$g_{n+1} = J_f(u_n) e^{f}_n - J_P(u_n) e^{P}_n$.
Substituting $J_f(u_n) = J_P(u_n) + \Delta_n$ yields:
\begin{equation*}
g_{n+1} = J_P(u_n) \, g_n + \Delta_n \, e^{f}_n .
\end{equation*}
Unrolling this linear recurrence from $g_0 = 0$ yields \eqref{eq:gap}.
\end{proof}

Equation~\eqref{eq:gap} shows that $g_n$ is driven solely by the Jacobian mismatch $\Delta_j$ acting on the current error $e^f_j$. We measure this directional mismatch along a unit vector $v$ via:
\begin{equation}
G_n(v) = \|\Delta_n v\|^2 .
\label{eq:G}
\end{equation}

\begin{corollary}[Optimal Probing Direction]
\label{cor:direction}
Let $\hat e_j = e^{f}_j / \|e^{f}_j\|$ be the unit error direction, $D$ the state dimension, $v$ a direction sampled uniformly from the unit sphere, and $\{q_1, \dots, q_D\}$ an orthonormal basis with $q_1 = \hat e_j$. For each $j < n$:
\begin{enumerate}
\item[(i)] The error generated at step $j$ satisfies $\|\Delta_j e^{f}_j\|^2 = \|e^{f}_j\|^2 \, G_j(\hat e_j)$.
\item[(ii)] The expected mismatch under random probing satisfies:
\begin{equation}
\mathbb{E}_v [G_j(v)] = \frac{1}{D} \, G_j(\hat e_j) + \frac{1}{D}\sum_{i \ge 2} G_j(q_i) .
\label{eq:expected}
\end{equation}
Hence, $G_j(\hat e_j) \le D \cdot \mathbb{E}_v [G_j(v)]$, with equality if and only if $G_j(q_i) = 0$ for all $i \ge 2$.
\end{enumerate}
\end{corollary}

\begin{proof}
(i) Write $\Delta_j e^{f}_j = \|e^{f}_j\| \, \Delta_j \hat e_j$ and take squared norms.
(ii) Expanding $v = \sum_i c_i q_i$ with $\mathbb{E}[c_i c_k] = \delta_{ik}/D$ yields $\mathbb{E}_v \|\Delta_j v\|^2 = \frac{1}{D}\sum_i \|\Delta_j q_i\|^2$. The bound follows from $G_j \ge 0$.
\end{proof}

\begin{remark}
\label{rem:direction}
By \cref{lem:gap} and \cref{cor:direction}(i), the contribution of step $j$ satisfies
\begin{equation*}
\|\Phi^{P}_{n,j+1} \Delta_j e^{f}_j\| \le \|\Phi^{P}_{n,j+1}\| \, \|e^{f}_j\| \, G_j(\hat e_j)^{1/2} .
\end{equation*}
Thus $g_n$ depends on the mismatch only through $G_j(\hat e_j)$. The remaining terms $\sum_{i \ge 2} G_j(q_i)$ in \eqref{eq:expected} play no role in driving the rollout error, yet take up $\frac{D-1}{D}$ of the penalty under random probing. Probing along $\hat e_j$ is therefore up to $D$ times more efficient than probing along random directions.
\end{remark}

\subsection{Error-Directed Jacobian Matching}
\label{sec:edjm}

\Cref{rem:direction} suggests penalising the mismatch along the error direction.
During fine-tuning we roll the current model out from a true state $u_0$ and take the unit direction of its rollout error, $\hat e_k = e_k/\|e_k\|$, which equals the direction of $e^f_k$ to first order.
Holding $\hat e_k$ fixed, error-directed Jacobian matching (\edjm{}) minimises
\begin{equation}
\mathcal{L} = \mathcal{L}_{\text{one-step}} + \mu\,\frac{1}{|K|}\sum_{k\in K}\frac{1}{D}\,G_k(\hat e_k)
= \mathcal{L}_{\text{one-step}} + \mu\,\frac{1}{|K|}\sum_{k\in K}\frac{1}{D}\big\|\big(J_f(u_k)-J_P(u_k)\big)\hat e_k\big\|^2 ,
\label{eq:own}
\end{equation}
where $K$ indexes the states of the rollout window and $\mu$ is the penalty weight.
Random-direction Jacobian matching (\rjm{}) replaces $\hat e_k$ by a random unit direction.
Both need only Jacobian-vector products of $f$ and $P$.

\subsection{When an Intervention Helps}
\label{sec:price}

Adding a Jacobian penalty introduces a trade-off: the model must balance one-step accuracy against trajectory stability. Throughout, the
amplification ratio $\rho$ and the error ratio $r$ are evaluated at a fixed
horizon $n$, which we suppress from the notation.

\begin{lemma}[Price of Intervention]
\label{lem:price}
Let an intervention replace $f$ by $f'$, change the amplification ratio from $\rho$ to $\rho'$, and scale every one-step defect by the same factor, $d'_m = \alpha\, d_m$.
The resulting rollout error ratio satisfies
\begin{equation}
r = \frac{\|e'_n\|}{\|e_n\|} = \frac{\rho'}{\rho}\,\alpha .
\label{eq:price}
\end{equation}
\end{lemma}

\begin{proof}
Unrolling \eqref{eq:counterfactual} from $e^{P}_0 = 0$ yields
$e^{P}_n = \sum_{m<n} \Phi^{P}_{n,m+1} d_m$, which is linear in $\{d_m\}$.
Scaling the defects to $d'_m = \alpha d_m$ gives
$\|e^{P\prime}_n\| = \alpha \|e^{P}_n\|$. Substituting into \eqref{eq:rho} yields
\begin{equation}
\frac{\|e'_n\|}{\|e_n\|}
= \frac{\rho' \|e^{P\prime}_n\|}{\rho \|e^{P}_n\|}
= \frac{\rho'}{\rho}\,\alpha . \qedhere
\end{equation}
\end{proof}

Uniform scaling of the defects is an idealisation. In our experiments we estimate $\alpha$ by the measured one-step cost $c$, the ratio of one-step errors after and before fine-tuning (\cref{eq:cost}).

\begin{proposition}[Floor, Threshold, and Local Minima]
\label{prop:threshold}
\mbox{}
\begin{enumerate}
\item[(i)] \emph{Floor:} Eliminating the mismatch completely
($\Delta_j = 0$ for $j < n$) yields $g_n = 0$ in \eqref{eq:gap}, so
$e^{f}_n = e^{P}_n$, $\rho' = 1$ to first order, and $r = \alpha/\rho$.
\item[(ii)] \emph{Threshold:} An intervention achieves an error
ratio $r < \tau$ if and only if
\begin{equation}
\rho > \rho^* = \frac{\rho'}{\tau}\,\alpha ,
\qquad \text{and in the floor case } (\rho' = 1),\quad \rho^* = \frac{\alpha}{\tau} .
\label{eq:threshold}
\end{equation}
\item[(iii)] \emph{Local minima:} If the initial parameters
lie at a local minimum of the one-step loss $\mathbb{E}\|d\|^2$, then
$\alpha \ge 1$ for any non-zero penalty weight, so
$\rho^* \ge \rho'/\tau$; in the floor case, $\rho^* \ge 1/\tau$.
\end{enumerate}
\end{proposition}

\begin{proof}
Claim (i) follows from \cref{lem:gap} with $\Delta_j = 0$, substituted into
\eqref{eq:price}. Claim (ii) is a rearrangement of \eqref{eq:price}.
For claim (iii), under uniform scaling $\alpha^2 = \mathbb{E}\|d'\|^2 / \mathbb{E}\|d\|^2$, and this is at least 1
because moving away from a local minimizer does not decrease the one-step loss.
\end{proof}

\begin{remark}
\label{rem:numbers}
Neither $\alpha$ nor $\rho'$ is known before retraining, so \eqref{eq:threshold} is not a decision rule by itself.
Under the approximation $\rho' \approx 1$, which our fine-tuned models satisfy (\cref{sec:res-ceiling}), it gives a necessary condition: starting from a local minimum of the one-step loss, no penalty weight achieves an error ratio below $\tau$ unless $\rho > 1/\tau$.
A starting model that is not at a local minimum can have $\alpha < 1$; we observe this for the 33k models (\cref{sec:res-ceiling}).
\end{remark}

\section{Experimental setup}
\label{sec:experiments}

Our experiments test contributions C2--C4; C1 is illustrated in \cref{sec:res-ledger}.
We answer these on small emulators (8--10k parameters) spanning five architectures and three PDEs, and repeat the analysis on larger emulators (about 33k parameters) on all four PDEs.
Throughout, errors are read at one Lyapunov time, before rollouts saturate, where the first-order decomposition of \cref{sec:method} is valid.
Full details are in \cref{app:systems,app:training,app:compute}.

\subsection{Systems and data}
\label{sec:systems}

We consider four one-dimensional chaotic PDEs with periodic boundaries: the Kuramoto--Sivashinsky equation (KS), the generalised KS equation (gKS), the complex Ginzburg--Landau equation in the phase-turbulent regime (CGLE), and the Nikolaevskiy equation.
Following APEBench~\citep{koehler2024apebench}, each system is discretised on 256 points and integrated with an ETDRK4 solver~\citep{cox2002etdrk4,kassam2005etdrk4} in double precision.
We implement the solver in JAX, so its Jacobian-vector products are exact.
For each system we generate 128 training trajectories of 50 steps, 16 validation trajectories of 100 steps and 30 test trajectories of 200 steps.
Each starts from a random initial condition that is first evolved onto the attractor.
Both studies use the same trajectories.
One Lyapunov time (LT), measured with the solver's tangent map, is 57, 123, 118 and 11 emulator steps for KS, gKS, CGLE and Nikolaevskiy.
Parameters and the data recipe are in \cref{app:systems}.

\subsection{Emulators}
\label{sec:emulators}

All emulators are residual one-step maps, $f(u)=u+\mathrm{net}(u)$, trained on the one-step mean squared error.
\Cref{tab:design} summarises the two studies.
The main study uses five architectures with 8.1--10.3k parameters (\cref{tab:app-arch}), which we refer to as 8--10k, on KS, gKS and CGLE.
We exclude the KS CNN, which diverges in every arm including the control.
The scaling check uses the APEBench ResNet and FNO presets at about 33k parameters on the four systems. We did not fine-tune the 33k FNO on Nikolaevskiy, which failed to learn the one-step map (relative one-step error of about 39\% on four of five seeds).
In both studies, fine-tuning starts from the pretraining checkpoint with the lowest validation rollout error.
Architectures and training schedules are in \cref{app:training}.

\begin{table}[t]
\centering\small
\begin{tabular}{@{}llccc@{}}
\toprule
 & systems & architectures & parameters & seeds \\
\midrule
Main study    & KS, gKS, CGLE       & CNN, ResNet, U-Net, FNO, Transformer & 8--10k & 3 \\
Scaling check & KS, gKS, CGLE, Nik. & ResNet, FNO                          & 33k    & 5$^{\dagger}$ \\
\bottomrule
\end{tabular}
\caption{Study design. The main study has 14 configurations and 42 seeds (KS CNN excluded); the scaling check has 7 configurations and 32 seeds. $^{\dagger}$Two seeds for gKS-ResNet, which showed no effect on both and was not extended.}
\label{tab:design}
\end{table}

\subsection{Fine-tuning methods}
\label{sec:arms}

All methods start from the same pretrained model, see the same training windows in the same order, and keep the one-step loss.
Training runs for 5k updates for the small models and 20k for the 33k models.

\paragraph{Error-directed Jacobian matching (\edjm{}).}
At each update we roll the model out over windows of 20 true states and take the unit direction of its error at each state, held fixed.
We then penalise the difference between the model's and the solver's Jacobian along that direction (\cref{eq:own}).
The penalty weight is fixed at the start so that the penalty equals 100 times the one-step loss.

\paragraph{Random-direction Jacobian matching (\rjm{}).}
The same penalty and calibration, applied along random unit directions at the same states.
It is an ablation that isolates the effect of the direction.

\paragraph{Baselines.}
Continued training trains on the one-step loss alone for the same budget, separating the effect of further training from that of Jacobian matching.
Input-noise injection (Noise)~\citep{sanchezgonzalez2020learning,stachenfeld2022learned} perturbs the inputs with Gaussian noise while keeping the clean target.
Its scale equals the starting model's RMS one-step error. We compare it with \edjm{} on the small models; one 33k configuration is reported in \cref{app:ledger}.

\subsection{Evaluation}
\label{sec:evaluation}

Each model is rolled out for 200 steps from the 30 test initial conditions.
For a model $f$ we write $E_f(n)$ for the relative error $\|e_n\|/\|u_n\|$ at step $n$, aggregated over trajectories by a geometric mean, and $n^\star$ for one Lyapunov time.
Our primary metric is the \emph{error ratio}
\begin{equation}
r = \frac{E_{\text{fine-tuned}}(n^\star)}{E_{\text{start}}(n^\star)},
\label{eq:ratio}
\end{equation}
and a fine-tuning run counts as an improvement when $r<0.9$.
The \emph{amplification ratio} $\rho\equiv\rho_{n^\star}$ (\cref{eq:rho}) is read at $n^\star$, for the starting model (the diagnostic) and after fine-tuning. For each seed it is the geometric mean over the 30 test trajectories of the per-trajectory ratio $\|e_{n^\star}\|/\|e^P_{n^\star}\|$; per-configuration values are medians over seeds.
The \emph{one-step cost} is
\begin{equation}
c = \frac{\varepsilon_{\text{fine-tuned}}}{\varepsilon_{\text{start}}},
\label{eq:cost}
\end{equation}
where $\varepsilon$ is the relative one-step error $\|d_n\|/\|u_n\|$, averaged by a geometric mean over all states of the test trajectories. It is our estimate of the defect scaling $\alpha$ in \cref{lem:price}.
Each run keeps its final checkpoint and its best checkpoint by validation rollout error, with the starting model as a candidate; each result states which it uses.
Comparisons between methods are paired by seed, with two-sided sign tests; ties, where both runs kept the starting model, are dropped from the tests.
All experiments ran on Kaggle T4 GPUs; compute is in \cref{app:compute}.

\section{Results}
\label{sec:results}

We evaluate model performance across two scale regimes: standard small emulators (8--10k parameters) followed by scaled configurations (33k parameters).

\subsection{The counterfactual error is a meaningful baseline (C1)}
\label{sec:res-ledger}

\begin{figure}[!htbp]
\centering
\includegraphics[width=\linewidth]{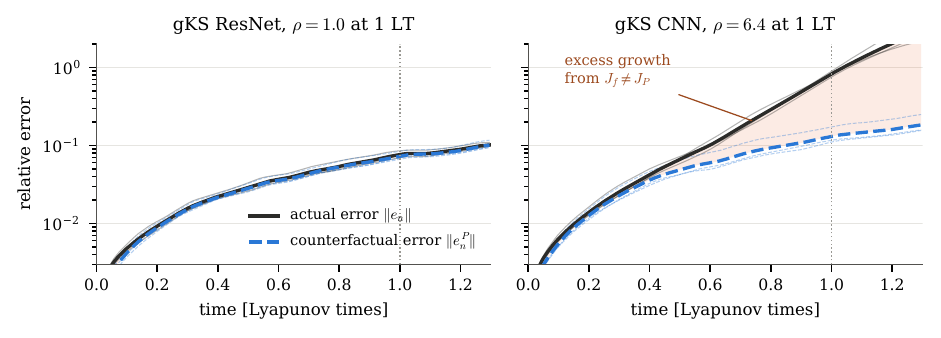}
\caption{Example on one system: actual error $\|e_n\|$ (solid) and counterfactual error $\|e^P_n\|$ (dashed) for two small emulators of gKS, one with $\rho\approx1$ and the configuration with the highest amplification ratio on this system. Thin lines: the three seeds; thick lines: their geometric mean over 30 test trajectories. Dotted: one Lyapunov time. Shaded: error in excess of the counterfactual. The $\rho$ in the panel titles is computed from the thick (seed-averaged) curves; \cref{tab:app-configs} reports the median over seeds (7.1 for the CNN). All configurations are shown in \cref{fig:app-ledger-small,fig:app-ledger-33k}.}
\label{fig:ledger}
\end{figure}

\Cref{fig:ledger} compares the actual error with the counterfactual error for two small gKS emulators.
For the ResNet the two coincide up to one Lyapunov time ($\rho=1.0$), which implies that the model propagates one-step errors like the true physics.
The CNN's counterfactual error at one Lyapunov time ($0.13$) is of the same order as the ResNet's ($0.07$), but its actual error ($0.84$) is six times larger ($\rho=6.4$ for the seed-averaged curves): the model amplifies its mistakes more than the true physics.
The gap between the curves is the part of the error that a better Jacobian can remove.
The same picture holds across all 21 configurations (\cref{fig:app-ledger-small,fig:app-ledger-33k}).

\FloatBarrier

\subsection{The amplification ratio identifies when fine-tuning helps (C2)}
\label{sec:res-diagnostic}

\begin{figure}[!htbp]
\centering
\includegraphics[width=\linewidth]{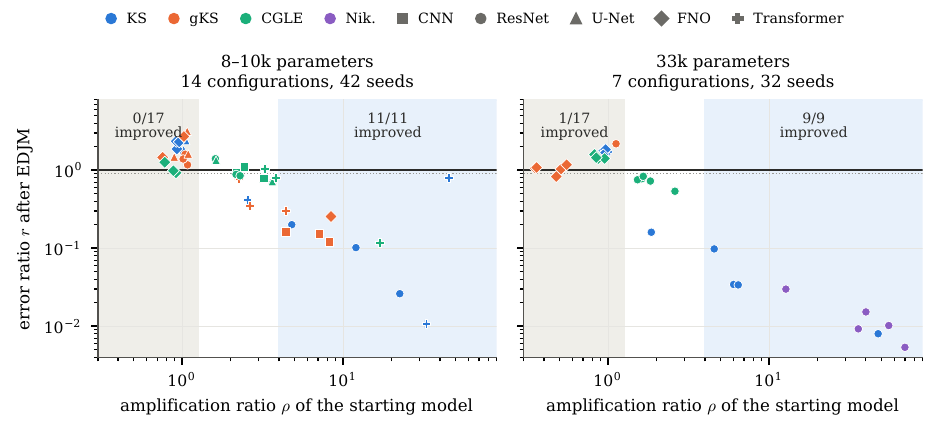}
\caption{Error ratio $r$ (\cref{eq:ratio}) after \edjm{} fine-tuning, final checkpoint, against the amplification ratio of the starting model. Each point is one seed; below the dotted line is an improvement of more than 10\%. Grey: $\rho\le1.25$; blue: $\rho\ge4$.}
\label{fig:diagnostic}
\end{figure}

\Cref{fig:diagnostic} plots the error ratio after \edjm{} fine-tuning (final checkpoint) against the amplification ratio of the starting model.
On the small models, none of the 17 seeds with $\rho\le1.25$ improved by more than 10\%, while all 11 with $\rho\ge4$ did.
The pattern holds at 33k: 1 of 17 seeds with $\rho\le1.25$ improved, against 9 of 9 with $\rho\ge4$. Hence, the amplification ratio identifies where Jacobian matching will help.
We chose the thresholds and the evaluation at one Lyapunov time after inspecting the results; the separation holds from half to one and a half Lyapunov times and for nearby thresholds (\cref{tab:robust}).

A low amplification ratio is a signal not to fine-tune for stability.
On the small models with $\rho\le1.25$, \edjm{} made the error worse, by a factor of $1.7$ in geometric mean over seeds, while continued training left it unchanged ($r=0.97$; final checkpoints; other methods in \cref{tab:app-bins}).
When $\rho\approx1$ the model already propagates its one-step mistakes as the true dynamics would, so there is no excess amplification to remove; the Jacobian penalty only raises the one-step error, and the rollout error rises with it (\cref{sec:res-ceiling}).
In this regime the remedy is a more accurate one-step model, not a better Jacobian.

\FloatBarrier

\subsection{The amplification ratio limits the gain (C3)}
\label{sec:res-ceiling}

\begin{figure}[tbp]
\centering
\includegraphics[width=\linewidth]{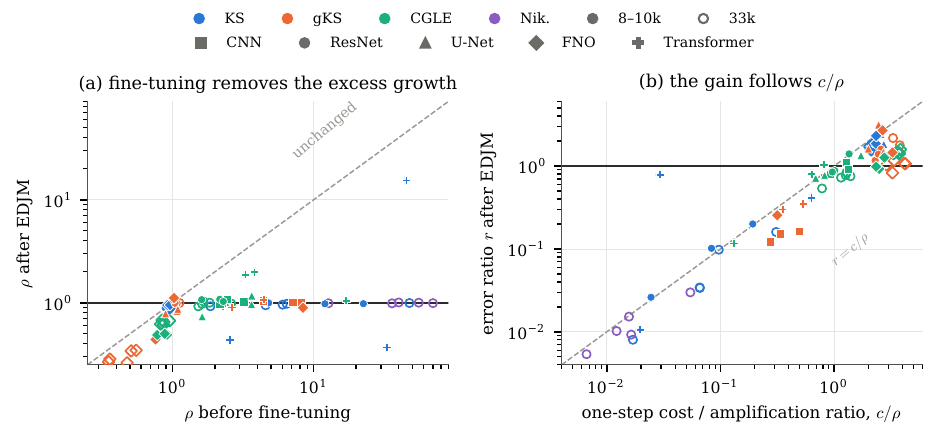}
\caption{(a) Amplification ratio before and after \edjm{} fine-tuning (dashed: unchanged). (b) Error ratio against $c/\rho$, where $c$ is the one-step cost (\cref{eq:cost}); dashed: $r=c/\rho$. Colour: system; marker shape: architecture; filled: 8--10k parameters; hollow: 33k. Final checkpoints, 74 seeds.}
\label{fig:ceiling}
\end{figure}

Fine-tuning reduces error amplification by bringing the model's error down to the counterfactual error.
After \edjm{}, the median amplification ratio reaches $0.99$ on the small models and $0.93$ at 33k, independent of its initial value (\cref{fig:ceiling}a; final checkpoints).

This is the floor case of \cref{prop:threshold}(i): with $\rho'\approx1$, \cref{lem:price} gives $r\approx\alpha/\rho$, and we estimate $\alpha$ by the one-step cost $c$.
On the small models with $\rho>1.25$ the one-step error grows by a median factor $c=2.2$.
At 33k it falls ($c=0.68$), which \cref{prop:threshold}(iii) excludes for a model at a minimum of the one-step loss.
Our starting checkpoints are selected by validation rollout error rather than by one-step loss, and ten of the 33k ResNets stopped at 145k of 150k pretraining updates, so these models were likely not at a minimum and fine-tuning could still lower their one-step error.
The measured error ratios track $c/\rho$ closely (\cref{fig:ceiling}b; log-space correlation $0.88$ on the small models and $0.99$ at 33k).
This also accounts for the mixed outcomes at intermediate $\rho$ in \cref{fig:diagnostic}: under our 10\% criterion, a model with $\rho=2$ improves only if fine-tuning raises its one-step error by less than a factor of $1.8$.
Because $c$ is known only after fine-tuning, this relation explains the size of the gain rather than predicting it; we return to this in \cref{sec:discussion}.

\subsection{Error-directed Jacobian matching (C4)}
\label{sec:res-arms}

\begin{figure}[!htbp]
\centering
\includegraphics[width=0.92\linewidth]{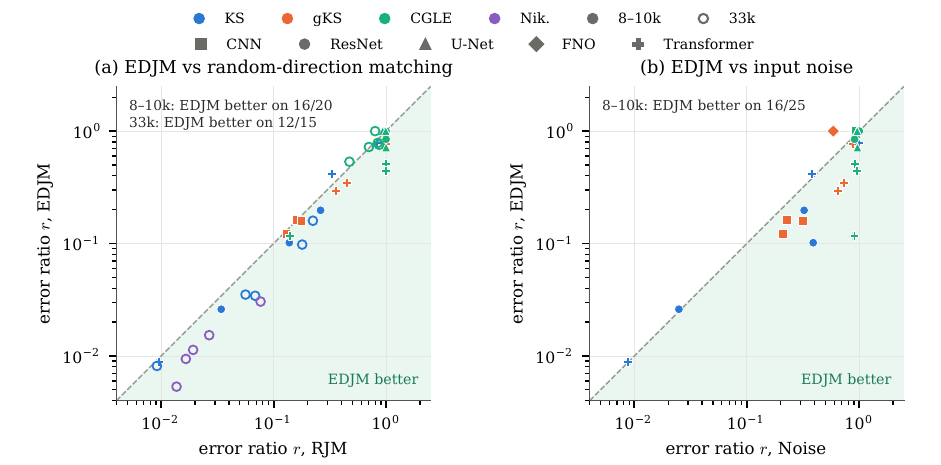}
\caption{Per-seed error ratio of \edjm{} against (a) \rjm{} and (b) Noise, on all seeds with $\rho>1.25$, validation-selected checkpoints. Points in the shaded region favour \edjm{}; points on the diagonal at $r=1$ are seeds where both runs kept the starting model. Colour: system; marker shape: architecture; filled: 8--10k; hollow: 33k. Panel (b) shows the small models only; the 33k comparison is in \cref{tab:arms}.}
\label{fig:arms}
\end{figure}

\Cref{cor:direction} predicts that the penalty is most effective along the direction of the error.
\Cref{fig:arms} compares \edjm{} with random-direction Jacobian matching (\rjm{}) on all seeds with $\rho>1.25$, and with input noise on the small models, using validation-selected checkpoints; full paired statistics, including continued training, are in \cref{tab:arms}.
\edjm{} beats \rjm{} on 16 of 20 small-model seeds ($p=0.012$) and on 12 of 15 seeds at 33k ($p=0.035$), with 14\% and 31\% lower error in geometric mean.
It also beats continued training (19 of 25 and 14 of 15), so the gain does not come from further optimisation.
Against input noise the evidence does not favour \edjm{}.
On the small models \edjm{} has lower error on 16 of 25 seeds, which is not significant ($p=0.23$).
A single comparison at 33k, on KS ResNet, is reported in \cref{app:ledger}.

\FloatBarrier

\section{Discussion}
\label{sec:discussion}

The amplification ratio is a cheap diagnostic of whether a model carries its errors forward like the true physics.
When $\rho\approx1$, Jacobian fine-tuning has nothing to remove and makes the error worse, and the effort is better spent on one-step accuracy.
When $\rho$ is large, it removes most of the excess growth.
The relation $r\approx c/\rho$ explains the size of the gain but does not predict it, since the one-step cost $c$ is known only after fine-tuning.

There are also some limitations.
The thresholds on $\rho$ and the evaluation at one Lyapunov time were chosen after inspecting the results, although the separation holds from half to one and a half Lyapunov times and for nearby thresholds (\cref{tab:robust}).
Input-noise injection matched \edjm{} on the small models (a single 33k comparison is in \cref{app:ledger}), so our evidence for \edjm{} is its advantage over random directions, not over other stabilisation methods.
The counterfactual requires solver Jacobian-vector products, which are available only for differentiable solvers and, in weather and climate, from tangent-linear models.
Our systems are one-dimensional, and future work would extend this analysis to two- and three-dimensional flows.

\section{Conclusion}
\label{sec:conclusion}

A rollout error in a chaotic system has two sources: the one-step mistakes of the emulator, amplified by the true dynamics, and the excess amplification of the emulator's own Jacobian.
Comparing the error with a counterfactual that propagates the same mistakes through the solver's Jacobian separates the two.
Across 21 configurations and two model sizes, the resulting amplification ratio identified when Jacobian fine-tuning helps: almost never when $\rho\le1.25$ (1 of 34 seeds), and always when $\rho\ge4$ (20 of 20).
Fine-tuning removes the excess, and the gain follows the ratio of the one-step cost to $\rho$.
Matching the solver's Jacobian along the model's own error direction outperformed random directions at the same cost.
Before stabilising an emulator, it is worth measuring how much of its error it is responsible for.

\section*{Code availability}
Code, starting checkpoints and per-run results, with scripts that regenerate every figure and table, are available at \url{https://github.com/manyagupta13/jacobian_matching}.

\bibliographystyle{plainnat}
\bibliography{references}

\appendix
\section{Systems and data}
\label{app:systems}

\paragraph{Equations.}
All systems are one-dimensional with periodic boundaries and are solved on $N=256$ points.
In the form we integrate,
\begin{align*}
\text{KS:}\quad & u_t + u u_x + u_{xx} + u_{xxxx} = 0, \\
\text{gKS:}\quad & u_t + u u_x + u_{xx} + \delta\, u_{xxx} + \nu\, u_{xxxx} = 0, \\
\text{CGLE:}\quad & A_t = A + (1 + i b_1) A_{xx} - (b_3 - i)\,|A|^2 A, \\
\text{Nikolaevskiy:}\quad & u_t + u u_x = -\partial_x^2\big[r_{\mathrm N} u - (1+\partial_x^2)^2 u\big].
\end{align*}
The CGLE field is complex and is represented as two real channels.
\Cref{tab:app-systems} lists the parameters, which follow~\citet{edson2019lyapunov,gotoda2015nonlinear,chate1994disordered,cox2007pattern}.

\begin{table}[!htbp]\centering\small
\begin{tabular}{@{}lcccccc@{}}
\toprule
system & parameters & $L$ & $\Delta t$ (substeps) & channels & $\lambda$ per step & 1 LT [steps] \\
\midrule
KS           & --                          & 60  & 0.25 (1) & 1 & 0.0175 & 57  \\
gKS          & $\delta=0.2$, $\nu=1$        & 200 & 0.25 (1) & 1 & 0.0081 & 123 \\
CGLE         & $b_1=3.5$, $b_3=1.4$         & 250 & 0.2 (4)  & 2 & 0.0085 & 118 \\
Nikolaevskiy & $r_{\mathrm N}=0.02$          & 200 & 10.0 (5) & 1 & 0.0918 & 11  \\
\bottomrule
\end{tabular}
\caption{Systems. $\Delta t$ is the time advanced by one emulator step, integrated with the given number of solver substeps. $r_{\mathrm N}$ is the supercriticality parameter $r$ of \citet{cox2007pattern}, renamed to avoid a clash with the error ratio.}
\label{tab:app-systems}
\end{table}

\paragraph{Solver.}
We use ETDRK4~\citep{cox2002etdrk4} with the contour-integral evaluation of~\citet{kassam2005etdrk4}, written in JAX in double precision.
The contour is a full circle of 64 points, because the half-circle form is valid only for real linear operators and the gKS and CGLE operators are complex.
The quadratic nonlinearities of KS, gKS and Nikolaevskiy are dealiased with the $2/3$ rule.
Solver Jacobian-vector products $J_P v$ are obtained by forward-mode differentiation through the solver.

\paragraph{Validation of the solver.}
Before training, each system passed four checks: tangent-map products agree with central finite differences (relative error below $10^{-9}$); the tangent map is linear to $10^{-14}$; the leading Lyapunov exponent is positive; and the grid resolves the linearly unstable band.

\paragraph{Data.}
An initial condition is $u_0(x)=\sum_{m=1}^{6}\xi_m m^{-1}\sin(2\pi m x/L+\phi_m)$ with $\xi_m\sim\mathcal N(0,1)$ and $\phi_m\sim\mathcal U(0,2\pi)$, scaled to a maximum amplitude of 0.5 (independently per channel draw for CGLE).
Each is evolved for 300 emulator steps (each of length $\Delta t$, \cref{tab:app-systems}) onto the attractor before recording.
For each system we generate 128 training trajectories of 50 steps, 16 validation trajectories of 100 steps and 30 test trajectories of 200 steps, with fixed random seeds (1000, 4242 and 773).
Both studies use the same trajectories, regenerated with identical settings and seeds.
Data are stored in single precision.

\paragraph{Lyapunov time.}
We propagate a perturbation of size $10^{-9}$ with the solver's tangent map for 200 steps from two attractor states, fit the slope of its log-norm over the middle 60\% of the run, and take the median.
One Lyapunov time is $1/\lambda$ emulator steps.

\section{Architectures and training}
\label{app:training}

\paragraph{Emulators.}
All emulators compute $f(u)=u+\mathrm{net}(u)$ with the last layer initialised to zero, and use periodic padding.
\Cref{tab:app-arch} lists the architectures.

\begin{table}[!htbp]\centering\small
\begin{tabular}{@{}llc@{}}
\toprule
architecture & configuration & parameters \\
\midrule
\multicolumn{3}{@{}l}{\emph{8--10k parameters (main study)}} \\
CNN         & 5 hidden convolutions, width 23, kernel 3, no internal skips & 8.2k \\
ResNet      & 8 residual blocks, width 13, kernel 3, ReLU & 8.4k \\
U-Net       & 3 levels, average-pool down, repeat up, skip concatenation & 10.3k \\
FNO         & 4 blocks, width 9, 12 modes, GELU (gKS: width 6, 32 modes) & 8.9k (9.9k) \\
Transformer & patches of 4 points, 2 pre-norm blocks, width 20, 2 heads & 8.1k \\
\midrule
\multicolumn{3}{@{}l}{\emph{33k parameters (APEBench presets)}} \\
ResNet      & \texttt{Res;26;8;relu}: 8 blocks, width 26 & 33.0k \\
FNO         & \texttt{FNO;12;18;4;gelu}: 4 blocks, width 18, 12 modes & 33.8k \\
\bottomrule
\end{tabular}
\caption{Architectures. Parameter counts are for one channel; the two-channel CGLE models are slightly larger. On gKS the small FNO uses 32 modes so that it covers the unstable band of the larger domain.}
\label{tab:app-arch}
\end{table}

\paragraph{Pretraining.}
All models are trained on the one-step mean squared error with Adam and gradient-norm clipping at 1.
The small models use 40k updates with batch size 64; the learning rate rises linearly from $10^{-4}$ to $10^{-3}$ over the first 5\% of updates and then decays to $10^{-5}$ with a cosine schedule.
The 33k models use 150k updates with batch size 32; the learning rate rises linearly to $10^{-3}$ over 2{,}000 updates and then decays to $10^{-5}$ with a cosine schedule.
Ten 33k ResNet runs stopped at 145k updates when the compute budget ran out.
In both studies the starting model for fine-tuning is the pretraining checkpoint with the lowest validation rollout error, defined under \emph{Checkpoints} below.

\paragraph{Fine-tuning.}
Each update draws 32 windows of 20 consecutive true states $u_0,\dots,u_{19}$, in the same order for every arm.
We use Adam with gradient-norm clipping at 1; the learning rate rises linearly to $10^{-4}$ over 300 updates and then decays to $10^{-6}$ with a cosine schedule, for 5k (small) or 20k (33k) updates.

\paragraph{Jacobian penalty.}
For \edjm{}, we roll the current model out from $u_0$ and take the unit direction of its error,
$\hat e_k = e_k/\|e_k\|$ with $e_k = f^{k}(u_0)-u_k$, for $k=1,\dots,19$; the direction is held fixed (no gradient), and states with a non-finite error are skipped.
For \rjm{}, $\hat e_k$ is replaced by a random unit Gaussian direction at the same states.
The loss is \cref{eq:own}, with the one-step mean squared error as $\mathcal{L}_{\text{one-step}}$ and $K=\{1,\dots,19\}$; $D$ is the number of state entries, and both Jacobian-vector products are computed by forward-mode differentiation, with the solver term treated as data.
The weight $\mu$ is set once, at update 0, so that the penalty equals 100 times the one-step loss, each averaged over 16 training batches.

\paragraph{Input-noise baseline (Noise).}
Inputs are perturbed as $u_k+\sigma\xi$ with $\xi\sim\mathcal N(0,I)$, and the target remains the clean solver step $u_{k+1}$.
The scale $\sigma$ is a multiple of the starting model's RMS one-step error, measured at update 0 over 16 training batches: $1\times$ for the small models, and $0.3\times$, $1\times$ and $3\times$ for the 33k KS-ResNet, of which we report $1\times$, the best.

\paragraph{Checkpoints.}
\label{app:checkpoints}
During fine-tuning we evaluate the validation rollout error 20 times per run for the small models and 40 times at 33k, and also at update 0.
The validation rollout error is the geometric mean over 100 steps of the relative error averaged over the 16 validation trajectories.
For gKS and CGLE these 100 steps are shorter than one Lyapunov time (123 and 118 steps), so checkpoint selection does not see the full evaluation horizon on these systems.
The \emph{best} checkpoint has the lowest validation rollout error, with the starting model as a candidate; the \emph{final} checkpoint is the last update.

\paragraph{Metrics.}
Relative errors $\|e_n\|/\|u_n\|$ are aggregated over the 30 test trajectories by a geometric mean, excluding non-finite values.
The counterfactual error $e^P_n$ propagates the model's measured one-step mistakes $d_n=f(u_n)-P(u_n)$ along the true test trajectory with the solver's Jacobian, starting from $e^P_0=0$; the model runs in single precision and the propagation in double precision.
The valid prediction time quoted in \cref{sec:introduction} is the first step at which the RMSE, normalised by the per-channel standard deviation of the training data, exceeds 0.5, taking the median over the test trajectories.

\section{Compute}
\label{app:compute}

All experiments ran on single NVIDIA T4 GPUs on Kaggle.
Pretraining the 33k models took 8.2 GPU-hours (FNO, 20 runs) and 10.0 GPU-hours (ResNet, 20 runs).
The small-model study, including pretraining and fine-tuning, took 3.3--3.9 GPU-hours per PDE.
Fine-tuning the 33k models took about 70 GPU-hours, including penalty variants not reported here, and the input-noise runs on the 33k KS-ResNet took 3.5 GPU-hours.

\section{Additional results}
\label{app:ledger}

\paragraph{Per-configuration results.}
\Cref{tab:app-configs} lists the amplification ratio and the outcome of \edjm{} fine-tuning for each configuration, the data behind \cref{fig:diagnostic}.

\begin{table}[!htbp]
\centering\small
\begin{tabular}{@{}lcccc@{}}
\toprule
configuration & seeds & $\rho$ & error ratio $r$ & improved \\
\midrule
\multicolumn{5}{@{}l}{\emph{8--10k parameters (14 configurations, 42 seeds)}} \\
CGLE FNO & 3 & 0.88 & 1 & 0/3 \\
KS FNO & 3 & 0.93 & 2.2 & 0/3 \\
KS U-Net & 3 & 1 & 2.3 & 0/3 \\
gKS FNO & 3 & 1 & 1 & 1/3 \\
gKS ResNet & 3 & 1 & 1.4 & 0/3 \\
gKS U-Net & 3 & 1.1 & 2 & 0/3 \\
CGLE ResNet & 3 & 2.2 & 1 & 2/3 \\
CGLE CNN & 3 & 2.4 & 0.93 & 1/3 \\
gKS Transformer & 3 & 2.6 & 0.43 & 3/3 \\
CGLE U-Net & 3 & 3.6 & 0.91 & 2/3 \\
CGLE Transformer & 3 & 3.8 & 0.46 & 2/3 \\
gKS CNN & 3 & 7.1 & 0.14 & 3/3 \\
KS ResNet & 3 & 12 & 0.081 & 3/3 \\
KS Transformer & 3 & 33 & 0.15 & 3/3 \\
\midrule
\multicolumn{5}{@{}l}{\emph{33k parameters (7 configurations, 32 seeds)}} \\
gKS FNO & 5 & 0.48 & 1 & 1/5 \\
CGLE FNO & 5 & 0.86 & 1.5 & 0/5 \\
KS FNO & 5 & 0.96 & 1.7 & 0/5 \\
gKS ResNet & 2 & 1.1 & 2 & 0/2 \\
CGLE ResNet & 5 & 1.7 & 0.72 & 5/5 \\
KS ResNet & 5 & 6 & 0.043 & 5/5 \\
Nik. ResNet & 5 & 40 & 0.012 & 5/5 \\
\bottomrule
\end{tabular}
\caption{Per configuration, ordered by $\rho$ within each size: median amplification ratio of the starting models, geometric-mean error ratio after \edjm{} (final checkpoint), and seeds with $r<0.9$.}
\label{tab:app-configs}
\end{table}

\paragraph{All methods by amplification ratio.}
\Cref{tab:app-bins} extends \cref{fig:diagnostic} to all fine-tuning methods.
On the small models with $\rho\le1.25$, no method improved more than one of 17 seeds, and both Jacobian methods increased the error on average.
Continued training and input noise at 33k are omitted: noise was run on one configuration only, and several continued-training rollouts diverge at 33k, which dominates their average.

\begin{table}[!htbp]
\centering\small
\begin{tabular}{@{}lccc@{}}
\toprule
 & $\rho\le1.25$ & $1.25<\rho<4$ & $\rho\ge4$ \\
\midrule
\multicolumn{4}{@{}l}{\emph{8--10k parameters (17 / 14 / 11 seeds)}} \\
\quad Cont.\ training & 0/17 (0.97) & 0/14 (1.08) & 0/11 (1.50) \\
\quad Noise           & 1/17 (0.95) & 3/14 (0.87) & 9/11 (0.25) \\
\quad \rjm{}          & 0/17 (1.71) & 5/14 (0.87) & 10/11 (0.18) \\
\quad \edjm{}         & 0/17 (1.73) & 9/14 (0.82) & 11/11 (0.13) \\
\midrule
\multicolumn{4}{@{}l}{\emph{33k parameters (17 / 6 / 9 seeds)}} \\
\quad \rjm{}          & 0/17 (2.40) & 6/6 (0.60) & 9/9 (0.034) \\
\quad \edjm{}         & 1/17 (1.43) & 6/6 (0.56) & 9/9 (0.018) \\
\bottomrule
\end{tabular}
\caption{Seeds improved by more than 10\% at one Lyapunov time, and in parentheses the geometric-mean error ratio over seeds, grouped by the amplification ratio of the starting model; final checkpoints.}
\label{tab:app-bins}
\end{table}

\paragraph{Paired comparisons.}
\Cref{tab:arms} gives the full paired statistics behind \cref{fig:arms}.

\begin{table}[!htbp]
\centering\small
\begin{tabular}{@{}llccc@{}}
\toprule
 & comparison & \edjm{} better & sign test $p$ & geo.\ mean ratio \\
\midrule
8--10k ($\rho>1.25$, 25 seeds) & vs \rjm{} & 16/20 & 0.012 & 0.86 \\
                                & vs Cont.\ training     & 19/25 & 0.015 & 0.36 \\
                                & vs Noise  & 16/25 & 0.23  & 0.72 \\
\midrule
33k ($\rho>1.25$, 15 seeds)    & vs \rjm{} & 12/15 & 0.035 & 0.69 \\
                                & vs Cont.\ training     & 14/15 & 0.001 & 0.07 \\
                                & vs Noise (KS ResNet) & 0/5 & 0.063 & 1.12 \\
\bottomrule
\end{tabular}
\caption{Paired comparisons of the error at one Lyapunov time, validation-selected checkpoints. Ties (both runs kept the starting model) are dropped from the counts and sign tests; the geometric-mean ratio of \edjm{} error to the other method's is over all seeds, ties included. Input noise at 33k uses the best of three noise levels.}
\label{tab:arms}
\end{table}

\paragraph{Input noise at 33k.}
We compared input noise with \edjm{} at 33k on one configuration, KS ResNet (last row of \cref{tab:arms}).
Input noise, tuned over three noise levels ($0.3\times$, $1\times$ and $3\times$ the RMS one-step error), was better at its best level, $1\times$, on all 5 seeds, by 12\% in geometric mean ($p=0.063$).
\edjm{} was run at a single setting and beat the other two noise levels on 5 and 4 of the 5 seeds, so the comparison favours noise in tuning effort; we did not tune \edjm{} or run noise on the other 33k configurations.

\paragraph{Sensitivity to the evaluation time and thresholds.}
The evaluation at one Lyapunov time and the thresholds $\rho\le1.25$ and $\rho\ge4$ were chosen after inspecting the results.
\Cref{tab:robust} repeats the analysis of \cref{fig:diagnostic} with both $\rho$ and the error ratio read at half and one and a half Lyapunov times.
Varying the low threshold over $\{1.1, 1.25, 1.5\}$ and the high threshold over $\{3, 4, 5\}$ at every evaluation time, at most 15.4\% (2 of 13) of the low-$\rho$ seeds and at least 87.5\% (14 of 16) of the high-$\rho$ seeds improved by more than 10\%.
The separation does not depend on these choices.

\begin{table}[!htbp]
\centering\small
\begin{tabular}{@{}lcccc@{}}
\toprule
 & \multicolumn{2}{c}{8--10k parameters} & \multicolumn{2}{c}{33k parameters} \\
\cmidrule(lr){2-3}\cmidrule(l){4-5}
evaluated at & $\rho\le1.25$ & $\rho\ge4$ & $\rho\le1.25$ & $\rho\ge4$ \\
\midrule
0.5 Lyapunov times & 1/18 & 7/7   & 0/17 & 7/7 \\
1 Lyapunov time    & 0/17 & 11/11 & 1/17 & 9/9 \\
1.5 Lyapunov times & 2/17 & 10/11 & 2/18 & 9/9 \\
\bottomrule
\end{tabular}
\caption{Seeds whose error improved by more than 10\% after \edjm{} fine-tuning (final checkpoint), with $\rho$ and the error ratio both read at the given time. The middle row is the analysis of the main text.}
\label{tab:robust}
\end{table}

\begin{figure}[!htbp]
\centering
\includegraphics[width=\linewidth]{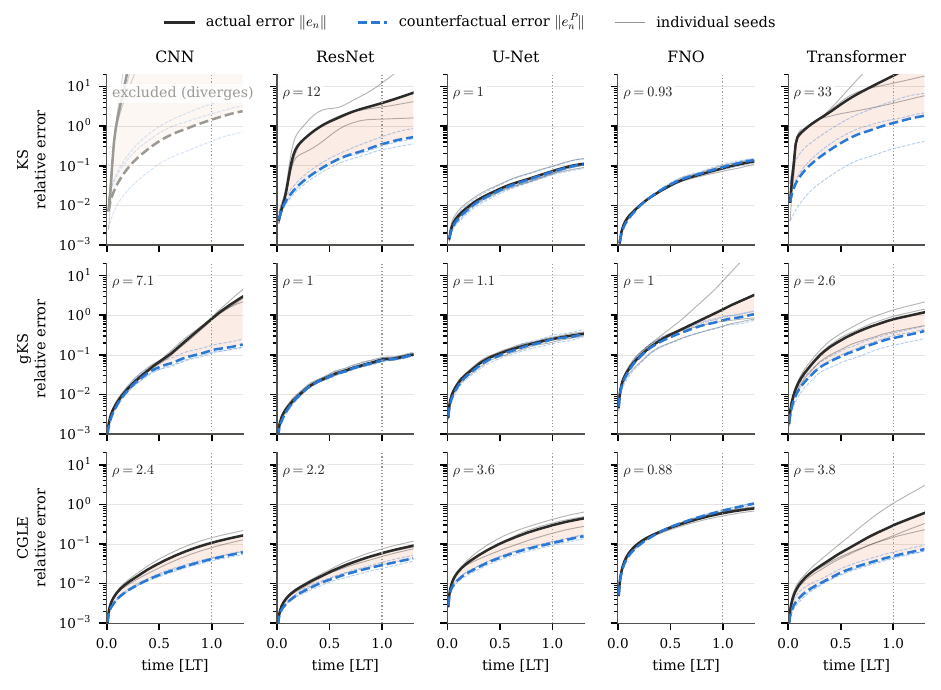}
\caption{Actual error (solid) and counterfactual error (dashed) for every small-model configuration; thin lines are the three seeds, thick lines their geometric mean over 30 test trajectories, and $\rho$ is the median over seeds at one Lyapunov time (dotted). The KS CNN, excluded from all analyses, is shown in grey.}
\label{fig:app-ledger-small}
\end{figure}

\begin{figure}[!htbp]
\centering
\includegraphics[width=\linewidth]{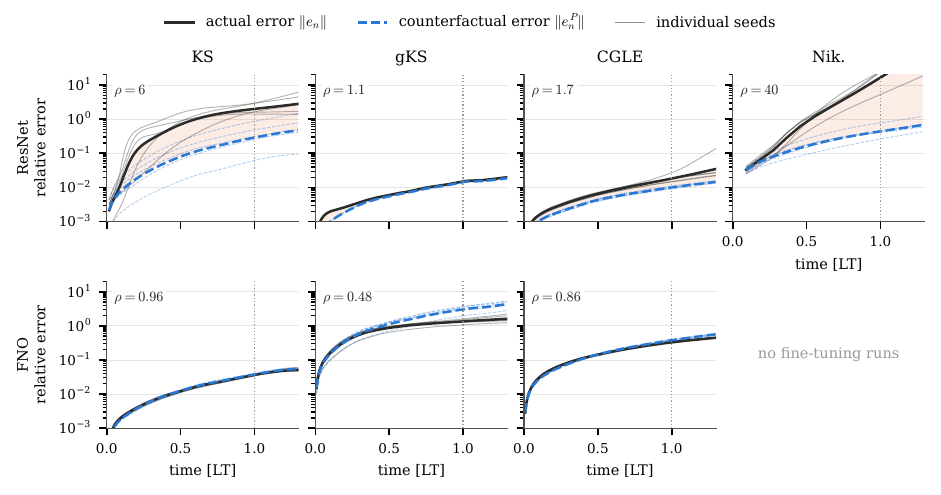}
\caption{As \cref{fig:app-ledger-small}, for the 33k-parameter models (five seeds; two for gKS ResNet).}
\label{fig:app-ledger-33k}
\end{figure}

\paragraph{The counterfactual error for every configuration.}
\Cref{fig:app-ledger-small,fig:app-ledger-33k} extend \cref{fig:ledger} to all configurations.
Where $\rho\approx1$ the actual error follows the counterfactual error up to one Lyapunov time; where $\rho$ is large it grows faster from the first steps.
Several FNOs have $\rho<1$, most clearly gKS FNO at 33k ($\rho=0.48$): there the model damps its own mistakes more than the solver would.

\clearpage

\end{document}